\documentclass{article} \usepackage{nips06submit,times}
\usepackage[dvips]{graphicx}

\title{Large-Margin kNN Classification Using a Deep Encoder Network}

\author{
Martin Renqiang Min\\
Department of Computer Science\\
University of Toronto\\
\texttt{minrq@cs.toronto.edu} \\
\AND
David A. Stanley \\
Electrical and Computer Engineering \\
University of Toronto \\
\texttt{davearthur.stanley@gmail.com} \\
\And
Zineng Yuan\\
Department of Molecular Genetics \\
University of Toronto\\
\texttt{zineng.yuan@utoronto.ca} \\
\And
Anthony Bonner \\
Department of Computer Science\\
University of Toronto \\
\texttt{bonner@cs.toronto.edu} \\
\And
Zhaolei Zhang \\
BBDMR \\
University of Toronto \\
\texttt{zhaolei.zhang@utoronto.ca} \\
}

\begin{document}

\maketitle

\begin{abstract}
KNN is one of the most popular classification methods, but it often fails to work well with inappropriate choice of distance metric or due to the presence of numerous class-irrelevant features. Linear feature transformation methods have been widely applied to extract class-relevant information to improve kNN classification, which is very limited in many applications. Kernels have been used to learn powerful non-linear feature transformations, but these methods fail to scale to large datasets. In this paper, we present a scalable non-linear feature mapping method based on a deep neural network pretrained with restricted boltzmann machines for improving kNN classification in a large-margin framework, which we call DNet-kNN. DNet-kNN can be used for both classification and for supervised dimensionality reduction. The  experimental results on two benchmark handwritten digit datasets show that DNet-kNN has much better performance than large-margin kNN using a linear mapping and kNN based on a deep autoencoder pretrained with retricted boltzmann machines.
\end{abstract}

\section{Introduction}
\label{intro}
kNN is one of the most popular classification methods due to its simplicity and reasonable effectiveness: it doesn't require fitting a model and it has been shown to have good performance for classifying many types of data. However, the good classification performance of kNN is highly dependent on the metric used 
for computing pairwise distances between data points. In practice, we often use Euclidean distances as similarity metric to calculate k nearest 
neighbors of data points of interest. To classify high-dimensional data in real applications, we often need to learn or choose a good distance metric. 

   Previous work on metric learning in \cite{Xing} and \cite{NCA} learns a global linear transformation matrix in the original feature space of data points to make similar data points stay closer while making dissimilar data points move farther apart using additional similarity or label information. In \cite{MLCC}, a global linear transformation is applied to the original feature space of data points to learn Mahalanobis metrics, which requires all data points in the same class collapse to one point. Making data points in the same class collapse to one point is unnecessary for kNN classification. It may produce poor performance when data points cannot be essentially collapsed to points,  which is often true for some class containing multiple patterns. An information-theoretic based approach is used to learn linear transformations in \cite{infoML}. In \cite{LMNN}, a global linear transformation is learned to directly improve kNN classification to achieve the goal of a large margin. This method has been shown to yield significant improvement over kNN classification, but the linear transformation often fails to give good performance in high-dimensional space and a pre-processing dimensionality reduction step by PCA is often required for success. 

   In many situations, a linear transformation is not powerful enough to capture the underlying class-specific data manifold; thus we need to resort to more powerful non-linear transformations, so that each data point will stay closer to its nearest neighbors having the same class as itself than to any other data in the non-linearly transformed feature space. Kernel tricks have been used to kernelize some of the above methods in order to improve kNN classification \cite{MLCC, infoML}. The method in \cite{LMCA} extends the work in \cite{LMNN} to perform linear dimensionality reduction to improve large-margin kNN classification and kernelized the method in \cite{LMNN}. However, the kernel-based approaches behave almost like template-based approaches. If the chosen kernel cannot well reflect the true class-related structure of the data, the resulting performance will be bad. Besides, kernel-based approaches often have difficulty in handling large datasets.

   We might want to achieve non-linear mappings by learning a directed multi-layer belief net or a deep autoencoder, and then perform kNN classification using the hidden distributed representations of the original input data. However, a multi-layer belief net often suffers from the "explaining away" effect, that is, the top hidden units become dependent conditional on the bottom visible units, which makes inference intractable; and learning a deep autoencoder with backpropagation is amost impossible because the gradient backpropagated to the lower layers from the output often becomes very noisy and meaningless. Fortunately, recent research has shown that training a deep generative model called Deep Belief Net is feasible by pretraining the deep net using a type of undirected graphical model called Restricted Boltzmann Machine (RBM) \cite{deepnet}.  RBMs produce "complementary priors" to make the inference process in a deep belief net much easier, and the deep net can be trained greedily layer by layer using the simple and efficient learning rule of RBM. The greedy layerwise pretraining strategy has made learning models with deep architures possible \cite{Hugo, Bengio}. Moreover, the greedy pretraining idea has also been successfully applied to initialize the weights of a deep autoencoder to learn a very powerful non-linear mapping for dimensionality reduction, which is illustrated in Fig. 1a) and 1b). Besides, the idea of deep learning has motivated researchers to use powerful generative models with deep architectures to learn better discriminative models \cite{deepembed}.

     In this paper, by combining the idea of deep learning and large-margin discriminative learning, we propose a new kNN classification and supervised dimensionality reduction method called DNet-kNN. It learns a non-linear feature transformation to directly achieve the goal of large-margin kNN classification, which is based on a Deep Encoder Network pretrained with RBMs as shown in Fig 2. Our approach is mainly inspired by the work in \cite{LMNN}, \cite{LMCA} and \cite{Hinton}. Given the labels of some or all training data, it allows us to learn a non-linear feature mapping to minimize the invasions to each data point's genuine neighborhood by other impostor nearest neighbors, which favours kNN classification directly. Previous researchers once used an autoencoder or a deep autoencoder for non-linear dimensionality reduction to improve kNN \cite{Hinton, nlnca}. None of these approaches used an objective function as direct as what we use here for improving kNN classification. The approach discussed in \cite{convolutioNNet} uses a convolution net to learn a similarity metric discriminatively, but it was handcrafted. Our approach based on general deep neural networks is more flexible and the connection weight matrices between layers are automatically learned from data.

     We applied DNet-kNN on the USPS and MNIST handwritten digit datasets for classification. The test error we obtained on the MNIST benchmark dataset is $0.94\%$, which is better than that obtained by deep belief net, deep autoencoder and SVM \cite{mnistSVM, Hinton, deepnet}. In addition, our fine-tuning process is very fast and converges to a good local minimum within several iterations of conjugate-gradient update. Our experimental results show that: (1) a good generative model can be used as a pretraining stage to improve discriminative learning; (2) pretraining with generative models in a layerwise greedy way makes it possible to learn a good discriminative model with deep architecture;  (3) pretraining with RBMs makes discriminative learning process much faster than that without pretraining; (4) pretraining helps to find a much better local minimum than without pretraining. These conclusions are consistent with the results of previous research trials on deep networks \cite{Hugo, Bengio, deepembed, deepnet, Hinton}.

     We organize this paper as follows: in section 2, we introduce kNN classification using linear transformations in a large-margin framework. 
In section 3, we describe previous work on RBM and training models with deep architectures. In section 4, we present DNet-kNN, which trains a Deep Encoder Network for improving large-margin kNN classification. In section 5, we present our experimental results on the USPS and MNIST \cite{mnist} handwritten digit datasets. In section 6, we conclude the paper with some discussions and propose possible extensions of our current method.

\section{Large-margin kNN classification using linear transformation}
 \label{lmnn}
In this section, we review the large-margin framework of kNN classification described in \cite{LMNN}. 
Given a set of data points $\mathcal{D} = \{{\mathbf x}^{(i)}, y_i: i = 1,\ldots, n\}$ and additional 
neighborhood information $\eta$, where ${\mathbf x}^{i} \in R^D$, $y_i \in \{1, \ldots, c\}$ for 
labeled data points, $c$ is the total number of classes, and $\eta_{il} = 1$ if $l$ is one of $i$'s $k$ target neighbors, we seek a distance function 
$d(i, j)$ for pairwise data points $i$ and $j$ such that the given neighborhood information 
will be preserved in the transformed feature space corresponding to the distance function. 
If $d(i, j)$ is based on Mahanalobis distances, then it admits the following form:
\begin{equation}\label{eq:Mahanalobis}
d_{\mathbf A} (i,  j) = ({\mathbf x}^{(i)} - {\mathbf x}^{(j)})^T {\mathbf A}^T {\mathbf A} ({\mathbf x}^{(i)} - {\mathbf x}^{(j)}),
\end{equation}
where ${\mathbf A}$ is a linear transformation matrix. Based on the goal of margin maximization, we learn the parameters 
of the distance function, ${\mathbf A}$, such that, for each data point $i$, the distance between $i$ and each data point 
$j$ from another class will be at least $1$ plus the largest distance between $i$ and its $k$ target 
neighbors. Using a binary matrix $y_{ij} = 1$ to represent that $i$ and $j$ are in the same class and $0$ otherwise for 
the labeled data points, we can formulate the above problem as an optimization problem:
\begin{equation}\label{eq:lmnn}
min_{\mathbf A}  \ \ \sum_{il}^{} \eta_{il} d_{\mathbf A} (i,  l) + C \sum_{ilj}^{}\eta_{il} (1-y_{ij}) h(1 + d_{\mathbf A} (i,  l) - d_{\mathbf A} (i,  j)), 
\end{equation}
where $C$ is a penalty coefficient penalizing constraint violations, and $h(\cdot)$ is a hinge loss function with $h(z) = max(z, 0)$. If ${\mathbf A}$ is a $D \times D$ matrix, this problem corresponds to the work in \cite{LMNN}; if ${\mathbf A}$ is a $d \times D$ matrix where
$d < D$, this problem corresponds to the work in \cite{LMCA}. When a non-square matrix ${\mathbf A}$ is learned for dimensionality 
reduction, the resulting problem is non-convex, stochastic gradient descent and conjugate gradient descent are often used to 
solve the problem. When ${\mathbf A}$ is constrained to be a full-rank square matrix, we can solve ${\mathbf A}^T {\mathbf A}$ directly 
and the resulting problem is convex. Alternating projection or simple gradient-based methods can be applied here \cite{LMNN}.

%

\begin{figure}[h]
\begin{center}
 \includegraphics[width =0.5\linewidth]{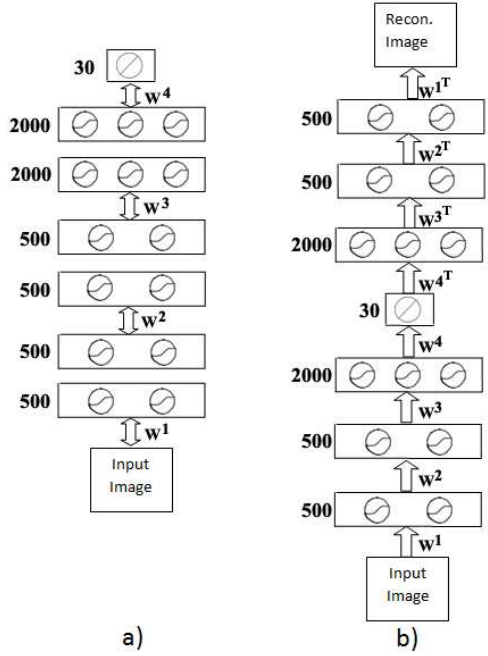}
\end{center}
\caption{RBM pretraining and deep autoencoder}
\end{figure}

\begin{figure}[h]
\begin{center}
 \includegraphics[width=0.5\linewidth]{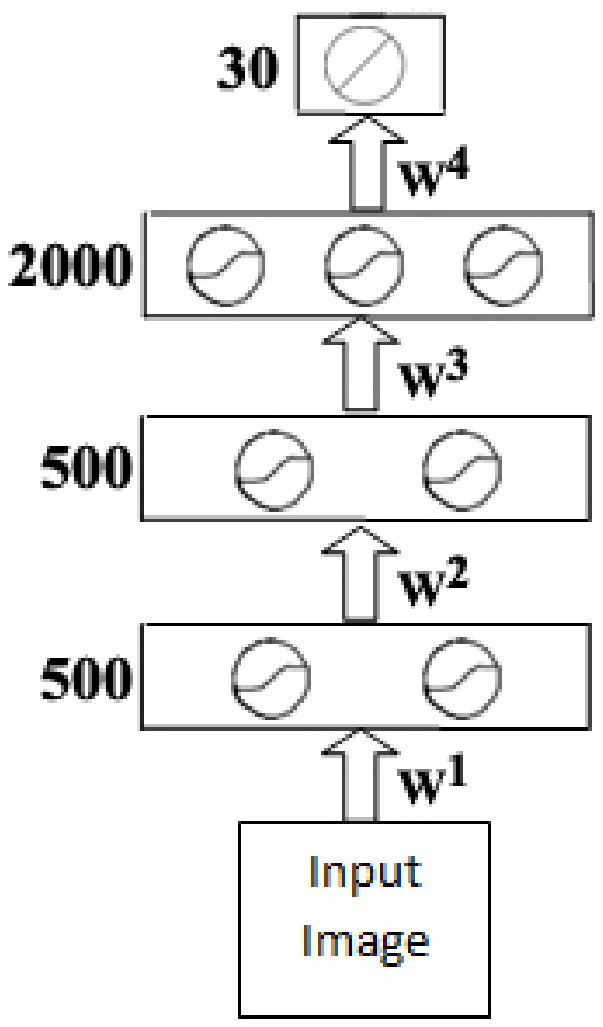}
\end{center}
\caption{Deep Encoder}
\end{figure}

\section{RBM and Deep Neural Network}
\label{nnet}
On large datasets, rich information existing in data features often enables us to build powerful generative models to learn the constraints and the structures underlying the given data. The learned information often reveals the characteristics of data points belonging to different classes.  In \cite{Hinton}, it is shown that a deep belief net composed of stacked Restricted Boltzmann Machines (RBM) can perform handwritten digit 
classification remarkably well \cite{RBM}.  RBM is an undirected graphical model with one visible layer ${\mathbf v}$ and one hidden layer ${\mathbf h}$. There are symmetric connections ${\mathbf W}$ between the hidden layer and the visible layer, but there are no within-layer connections. For a RBM with stochastic binary visible units ${\mathbf v}$ and stochastic binary hidden units ${\mathbf h}$, the joint probability distribution of a configuration $({\mathbf v}$, ${\mathbf h})$ of RBM is defined based on its energy as follows:
\begin{eqnarray}
E({\mathbf v}, {\mathbf h}) = -\sum_{ij} W_{ij}v_ih_j - \sum_i v_ib_i -\sum_j h_jc_j \\
p(v, h) = \frac{1}{Z} exp(-E({\mathbf v}, {\mathbf h})),
\end{eqnarray}
where ${\mathbf b}$ and ${\mathbf c}$ are biases, and $Z$ is the partition function with $Z = \sum_{{\mathbf u}, {\mathbf g}} exp(-E({\mathbf u}, {\mathbf g}))$.
The good property due to the structure of RBM is that, given the visible states, each hidden unit is conditionally independent, and given the 
hidden states, the visible units are conditionally independent. 
\begin{eqnarray}
p(v_i = 1|{\mathbf h}) = \sigma(\sum_{j} W_{ij}h_j + b_i),\\
p(h_j = 1|{\mathbf h}) = \sigma(\sum_{i} W_{ij}v_i + c_j),
\end{eqnarray}
where $\sigma(z) = \frac{1}{1 + exp(-z)}$. This beneficial property allows us to get unbiased samples from the posterior distribution of the hidden units given an input data vector. By minimizing the negative 
log-likelihood of the observed input data vectors using gradient descent, the update rule for the weight ${\mathbf W}$ turns out to be, 
\begin{equation}
\Delta W_{ij} = \epsilon (<x_j h_j>_{data} - <x_i h_j>_{\infty})
\end{equation}
where $\epsilon$ is learning rate, $<\cdot>_{data}$ denotes the expectation with respect to the data distribution and $<\cdot>_{\infty}$ denotes the expectation with respect to the model distribution. In practice, we do not have to sample from the equilibrium distribution of the model, and even one-step reconstruction samples work 
very well \cite{PoE}.  
\begin{equation}
\Delta W_{ij} = \epsilon (<x_j h_j>_{data} - <x_i h_j>_{1})
\end{equation}
Although the above update rule does not follow the gradient of the log-likelihood of data exactly, it approximately follows the gradient of another objective function \cite{Carreira}. In \cite{deepnet}, it is shown that a deep belief net based on stacked RBMs can be trained greedily layer by layer. Given some observed input data, we train a RBM to get the hidden representations of the data. We can view the learned hidden representations as new data and train another RBM. We can repeat this procedure many times. It is shown that in \cite{deepnet}, under this greedy training strategy, we always get better hidden representations of the original input data if the number of features in the added layer does not decrease, and a varational lower bound of the log-likelihood of the observed input data never decreases. In \cite{Hinton}, the greedy training strategy is used to initialize the weights of a deep autoencoder as shown in Fig. 1a) and then back-propagation is used for tuning the weights of the network as shown in Fig 1b). This time the lower-bound guarantee no longer holds, but the greedy pre-training still  works very well in practice \cite{Hinton}.

\section{Large-margin kNN classification using deep neural networks}
\label{largemargin}
     The work in \cite{Hinton} and \cite{deepnet} made full use of the capabilities of generative models, but label information is only weakly used. In the following, we describe DNet-kNN, in which we use stacked RBMs to initialize the weights of an encoder with 4 hidden layers as shown in Fig. 2. Then we fine-tune the weights of the encoder by minimizing the following objective function:
\begin{equation}
\ell_{ilj} = h(1 +  d_f(i, l) - d_f(i, j)),
\end{equation}
\begin{equation}\label{eq:lmnn-no-nn}
min_f \ \ \ell_f  =  \sum_{ilj}^{}  \eta_{il} \gamma_{ij} \ell_{ilj},
\end{equation}
where $\gamma_{ij} = 1$ if and only if $i$ is an impostor neighbour of $j$, which will be discussed in details later. The definition of $\eta_{il}$ is the same as discussed before in section \ref{lmnn}, and $d_f(i, j) = ||f({\mathbf x}^{(i)}) -  f({\mathbf x}^{(j)})||^2$. The function $f(\cdot):  R^D \rightarrow R^d$ is a continuous non-linear mapping, and each component of $f({\mathbf x})$ is a continuous function of the input vector ${\mathbf x}$, and the parameters ${\mathbf W}$ of $f$ are connection weight matrices in a deep neural  network. For example, in Fig. 1, ${\mathbf W} = \{ {\mathbf W}^{(i)}, i = 1, \ldots, 4\}$. This equation differs from Eqn 2 in two ways. First, the distance between data points $i$ and $j$ is computed by the Euclidean distance using the feature vectors output by the top layer of the encoder $f$ in Fig. 1. Secondly, the objective function $\ell_f$  focuses on maximizing the margin and neglects the term reducing the distance between nearest neighbors. Additionally, unlike Hinton's deep autoencoder, we no longer minimize the reconstruction error, since it was found that this criterion reduced the ability of the code vectors to accurately describe subtle differences between classes in practice.

      To reduce the complexity of the backpropagation training, we use simplified versions of $\eta_{il}$ and $\gamma_{ij}$ in the objective function \ref{eq:lmnn-no-nn}, as compared to those described in Section 2. For each index $i$, $\eta_{il} = 1$ only if $l$ is one of $i$'s top $k$ nearest neighbors among the data points having the same class label as $i$ (inclass nearest neighbors). In contrast, for each $i$, the $j$s for which $\gamma_{ij} = 1$ are selected from the set of impostor nearest neighbors of $i$, which is the union of the $m$ nearest neighbors from each and every class other than the class of $i$. For example, in the case of digit recognition with ten classes, there are a total of $m \times 9$ impostor $j$s for each data point $i$. This method of choosing impostor nearest neighbors is optimal for kNN classification because, by selecting $m$ impostor neighbors from every other class, we help ensure that all potential competitors are removed.
   
Let ${\mathbf y}^{(i)} = f({\mathbf x}^{(i)})$ be the low dimensional code vector of ${\mathbf x}^{(i)}$ generated by the Deep Encoder Network.
      Then the time complexity of computing Eq. \ref{eq:lmnn-no-nn} is $O((c-1)kmn)$, which is a significant improvement over the time complexity of Eq. \ref{eq:lmnn}, which is $O(kn^2)$. For the purposes of calculating nearest neighbors and impostor nearest neighbors, we use Euclidean distances in the pixel space. This means that the $\eta_{il}$ and $y_{ij}$ do not need to be recalculated each time the code vectors are updated. Unfortunately, due to the non-linear mapping, this may mean that ordinary data points in the pixel space may become impostors in the code space and will not be taken into account in the objective function. However, it is likely that the mapping is quasi-linear. Therefore, by taking a large value for $m$, we find that this captures most of the impostors in the code space, as evidenced by our low kNN classification errors. In our experiments, we use $k$=5 and $m=30$.

			To improve the computation time of calculating the objective function gradient, a $((c-1)kmn)$ x $3$ matrix of triples was generated. These triples represent the sets of all allowed indices $i$, $j$, and $l$ in Eq \ref{eq:lmnn-no-nn} for which $\eta_{il}$ and $\gamma_{ij}$ are non-zero. Therefore, in the triples matrix, the entries in the 2nd column represent the inclass nearest neighbors relative to the first column, and the entries in the 3rd column represent the impostor nearest neighbors relateive to the first column. The triples matrix is used in calculating both the gradient of the objective function, and the value of the objective function itself.


			The gradient of the objective function, relative to the code vector ${\mathbf y}_{i}$ = $f({\mathbf x}^{(i)})$ is given by:
\begin{equation}\label{eq:grad}
\frac{\partial \ell_f}{\partial {\mathbf y}_{i}} = -2\sum_{jl}^{}\eta_{il}\gamma_{ij}\theta_{ilj}({\mathbf y}_{l}-{\mathbf y}_{j}) - 2\sum_{jk}^{}\eta_{ki}\gamma_{kj}\theta_{kij}({\mathbf y}_{k}-{\mathbf y}_{i}) + 2\sum_{kl}^{}\eta_{kl}\gamma_{ki}\theta_{kli} ({\mathbf y}_{k}-{\mathbf y}_{i}) 
\end{equation}
 where $\theta$ is the flag for margin violations: $\theta_{klj} = 1 $ if $  (1 +  d_f(k, l) - d_f(k, j)) > 0$.

			While this equation is unwieldy for Matlab implementation, the use of the triples matrix makes the computation much easier. In Eq \ref{eq:grad}, we calculate each sum individually, using the triples matrix to determine the appropriate indicies, and then combined them later. For example, to determine the value of the first summation term,
\begin{equation}
\sum_{jl}^{}\eta_{il}\gamma_{ij}\theta_{ilj}({\mathbf y}_{l}-{\mathbf y}_{j}),\nonumber\\
\end{equation}
we simply search through the triples matrix to identify all the triples that yield a margin violation ($\theta=1$). Then, we choose those that have index $i$ in their first column. Thus, this specific set of triples tells us the appropriate indices to use in the first sum. Specifically, the second column of the triples matrix becomes the l index values, and the third column becomes the j index values. Likewise, the same strategy is repeated for the second and third summations.

\section{Experimental Results}

We will test our model DNet-kNN for both classification and dimensionality reduction on two handwritten digit dataset: USPS and MNIST.
We demonstrate two different types of classification: standard kNN and minimum energy classification. For standard kNN, after we finish learning the non-linear mapping by discriminative fine-tuning, we can directly compute pairwise Euclidean disitances for kNN classication, which is used in DNet-kNN. Alternatively, for minimum energy classification, after we calculate the feature vectors of training data and test data, we can also predict the class label of a test data point by the class to which the test data point is assigned to have the lowest energy defined by Eq. \ref{eq:lmnn-no-nn}. This minimum energy classification is denoted by "-E" in the experimental results. In both USPS and MINIST experiments, we set $k = 5$ and $m = 30$.

\subsection{Experimental Results on USPS Dataset}

We downloaded the USPS digit dataset from a public link\footnote{http://www.cs.toronto.edu/~roweis/data/usps\_all.mat}. From this dataset, several different preparations are used. The first preparation is USPS-fixed, which takes the first 800 data points from each of the ten digit classes to create an 8000 point training set. The test set for USPS-fixed then consists of a further 3000 data points, with 300 from each data class.

Second to sixth preparations, called USPS-random1 to USPS-random5 are then obtained from USPS-fixed by randomly shuffling the data points for each class between training and testing datasets.


\begin{figure}[h]
\begin{center}
\includegraphics[width=4.5in]{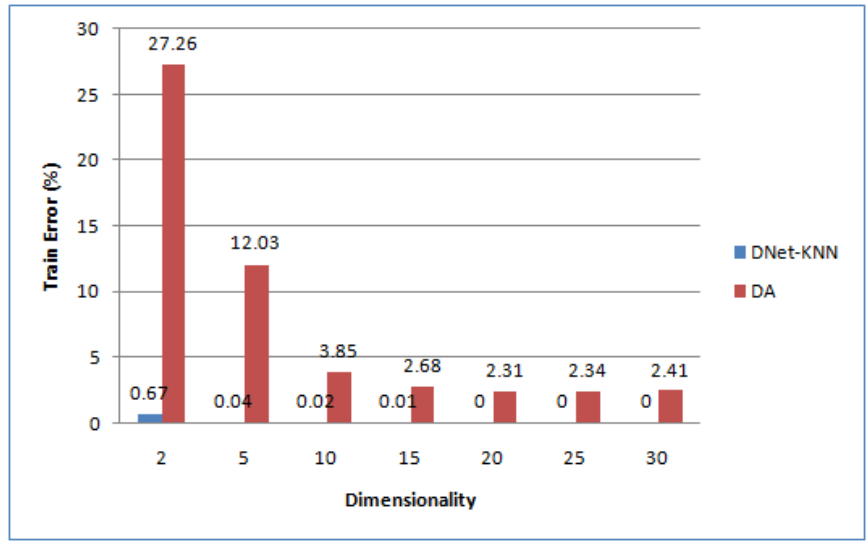} 
\end{center}
\caption{Training error on USPS-fixed using two different classification methods: the Deep Neural Network kNN classifier (DNet-kNN) and the deep autoencoder (DA).}
\end{figure}

\begin{figure}[h]
\begin{center}
\includegraphics[width=4.5in]{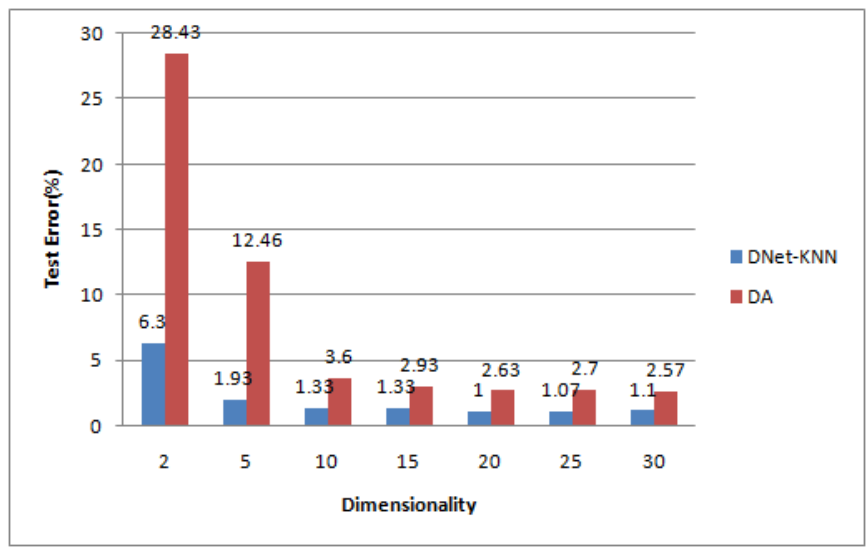}  
\end{center}
\caption{Test error on USPS-fixed using two different classification methods: the Deep Neural Network kNN classifier (DNet-kNN) and the deep autoencoder (DA).}
\end{figure}

In Figures 3 and 4, we observe the training errors and test errors for different dimensionality codes. In all cases, the DNet-kNN classification outperforms the deep autoencoder (DA). Furthermore, as the dimensionality of the codes increases, the classification accuracy increases. This trend coninues from d=2 up till d=15, and then levels off.
%

\begin{figure}[h]
\begin{center}

\includegraphics[width=6.5in]{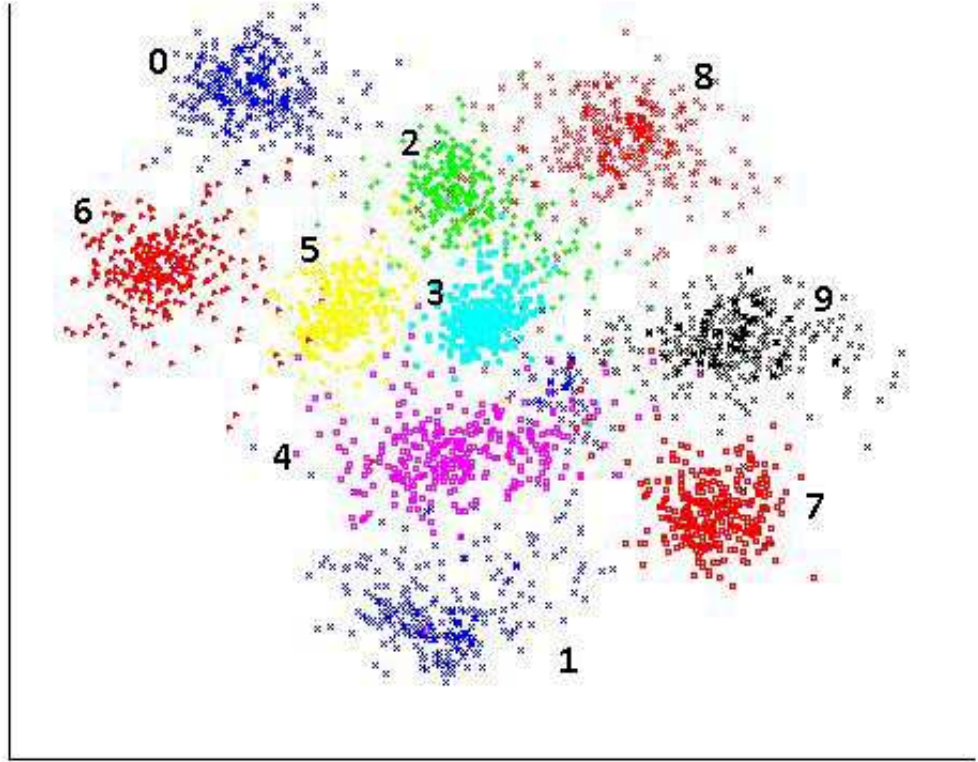} 
\end{center}
\caption{Two-dimesional embedding of USPS-fixed test data using the Deep Neural Network kNN classifier (DNet-kNN).}
\end{figure}

\begin{figure}[h]
\begin{center}
\includegraphics[width=6.5in]{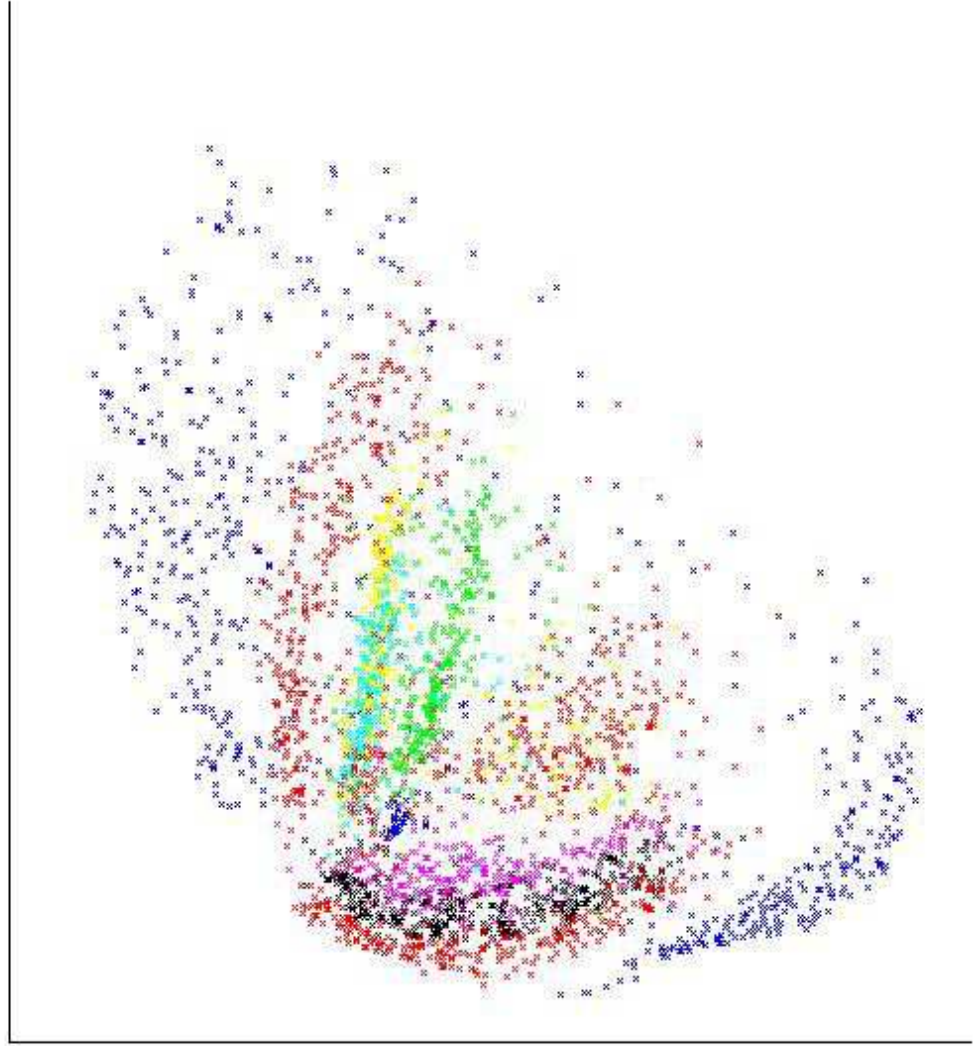}
\end{center}
\caption{Two-dimesional embedding of USPS-fixed test data using the deep autoencoder (DA).}
\end{figure}

Figures 5 and 6 compare DNet-kNN 2D dimensionality reduction with the deep autoencoder. The DNet-kNN clearly produces superior clustering of data point classes in two-dimensional space. There are still some class overlaps, however, because the backpropagation algorithm we use to optimize for kNN classification is not the best choice to improve visualization. This is because the objective function chooses which data points it considers to be in the set of impostor nearest neighbours (allowed $j$'s) using the pixel space rather than the code space (see Section \ref{largemargin}). However, visualization requires reduction to very low-dimensional spaces, and the mapping from pixel space to code space must become highly non-linear as dimensionality is reduced. Therefore, the pixel space becomes a poorer representation of spatial relationships in the code space and the correct choice of impostor nearest neighbours becomes less reliable during visualization.

\begin{table}
\caption{Training error  of  different methods on 5 random splits of USPS dataset (\%). The lowest errors are shown in bold. DNet code dim=30}
\label{tab:train}
\begin{center}
\begin{tabular}{l|c|c|c|c}
    Dataset       & kNN	& DA 	& LMNN	& DNet-kNN\\
\hline
USPS-random1	&5.12 &	2.66	&0.76	& \textbf{0.00}\\
USPS-random2	&5.09	&2.35	&1.10	&\textbf{0.00}\\
USPS-random3	&4.95	&2.28	&0.71	&\textbf{0.00}\\
USPS-random4	&5.08	&2.24	&0.85	&\textbf{0.00}\\
USPS-random5	&4.93	&2.48	&0.95	&\textbf{0.01}\\

\end{tabular}
\end{center}
\end{table}

\begin{table}
\caption{Test error of  different methods on the same 5 random splits of USPS dataset. "-E" denotes the  energy classification 
method (\%). The lowest errors are shown in bold. DNet code dim=30}
\label{tab:train}
\begin{center}
\begin{tabular}{l|c|c|c|c|c|c}
Dataset	 &kNN	 &DA	 & LMNN	& LMNN-E	  & DNet-kNN	   &DNet-kNN-E\\
\hline
USPS-random1	 &4.47  &2.80	 &2.20	 &1.77	 &\textbf{1.20}	 &1.43\\
USPS-random2	 &4.93	 &2.36	 &2.13	 &1.53	 &\textbf{0.87}	 &0.97\\
USPS-random3	 &5.23	 &2.33	 &2.36	 &1.80	 &\textbf{1.43}	 &1.50\\
USPS-random4	 &4.17	 &1.93	 &2.33	 &1.80	 &\textbf{1.06}	 &1.20\\
USPS-random5	 &5.37	 &2.23	 &1.93	 &1.63	 &1.13	 &\textbf{1.00}\\
\end{tabular}
\end{center}
\end{table}

Tables 1 and 2 show respectively the training and test error on multiple USPS-random data sets. DNet-kNN almost consistently outperforms the other methods.

\subsection{Experimental Results on MNIST Dataset}

This section deals with the MNIST dataset, which is another digit set available online.\footnote{http://yann.lecun.com/exdb/mnist/} This dataset contains 60,000 training samples and 10,000 test samples. For the USPS dataset, it was possible to do both the pre-training and the backpropagation on a single batch of data. However, given the size of the MNIST dataset, the training data had to be broken into smaller batches of 10,000 randomly selected datapoints. Then, RBM training and backpropagation could be applied iteratively to each batch. In all our experiments, batch size was set to 10,000.



\begin{figure}[h]
\begin{center}
\includegraphics[width=7in]{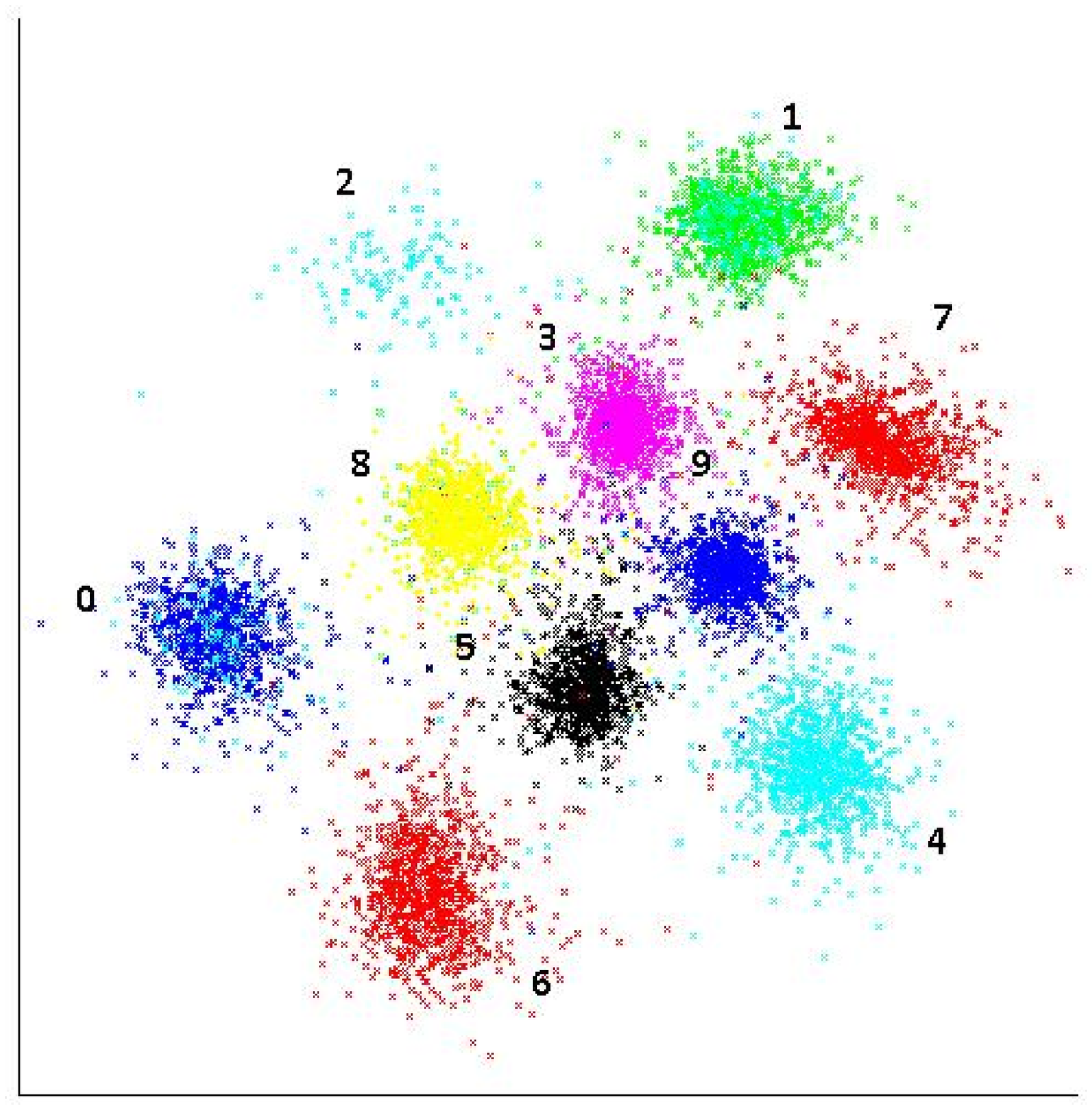} 
\end{center}
\caption{Two-dimesional embedding of 10,000 MNIST test data using the Deep Neural Network kNN classifier (DNet-kNN).}
\end{figure}

\begin{figure}[h]
\begin{center}
\includegraphics[width=7in]{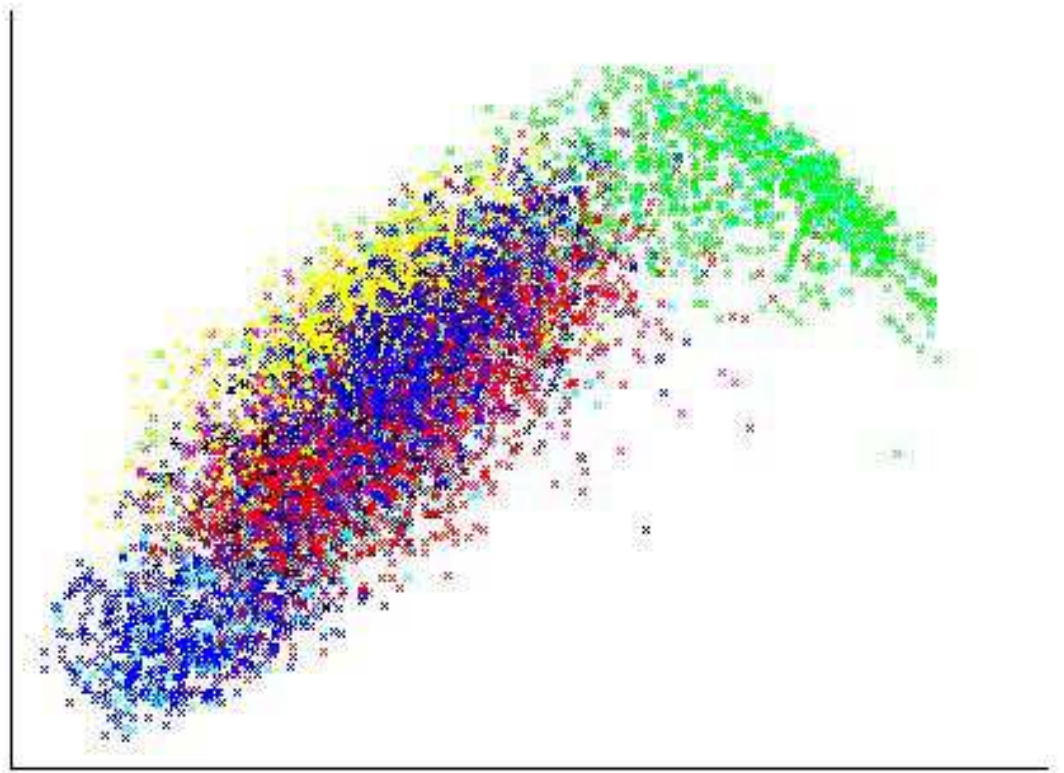}
\end{center}
\caption{Two-dimesional embedding of 10,000 MNIST test data using the deep autoencoder (DA).}
\end{figure}

Figure 7 and 8 show the mapping of the MNIST dataset onto a reduced space using the DNet-kNN and the deep autoencoder. As with the USPS dataset, it shows a significant improvement above the deep autoencoder.

\begin{table}
\caption{Test error of different methods on the benchmark MNIST dataset. "-E" denotes the  energy classification 
method (\%). For different kNN-based methods, k = 5.}
\label{tab:train}
\begin{center}
\begin{tabular}{l|l}
Methods	 & results\\
\hline
DNet-kNN (dim = 30, batch size = 10k) & \textbf{0.94}\\
DNet-kNN-E (dim = 30, batch size = 10k) & 0.95 \\
Non-linear NCA  based on a Deep Autoencoder (\cite{nlnca} & 1.03 \\
Deep Belief Net \cite{deepnet} & 1.25 \\
SVM: degree 9 \cite{mnistSVM} & 1.4 \\
kNN  & 3.05 \\
LMNN (dim = 30) & 2.62\\
LMNN-E (dim = 30) & 1.58 \\

\end{tabular}
\end{center}
\end{table}

This final table shows the classification error of the DNet-kNN as compared to other common classification techniques on the MNIST dataset. Despite the fact that we must use batches, the DNet-kNN still produces the best classifications. This indicates that the DNet-kNN classifier is highly robust, since it can perform well when limited to seeing only part of the dataset at any one time.

Finally, it is worth noting that, unlike the deep autoencoder, the fine tuning of the DNet-kNN classifier during backpropagation displays extremely fast convergence. Often, the error reaches a minimum after three to five epochs. This is due to the fact that the RBM pretraining has provided an ideal starting point and also that we are using a supervised learning algorithm, as opposed to an unsupervised algorithm as in the deep autoencoder.
\section{Discussions and future research}
In this paper, we present a new non-linear feature mapping method called DNet-kNN that uses a deep encoder network pretrained with RBMs to 
achieve the goal of large-margin kNN classification. Our experimental resuls on USPS and MNIST handwritten digits show 
that DNet-kNN is powerful in both classification and non-linear embedding. Our results suggest that, pretraining with a good generative model is very helpful for learning a good discriminative model, and the pretraining makes discriminative learning much faster, and it often help it find a much better local minimum especially in a deep architecture than without pretraining. Our findings are consistent as the idea discussed in \cite{Hinton222}.
   
    On huge dataset, the current implemention of our method only works by using mini-batches. We essentially compute the genuine nearest neighbors and impostor nearest neighbors in each mini-batch, which might be not optimum over the whole dataset. In the future, we plan to develop a dynamic version, in which the mini-batches will change dynamically during training and we dynamically update the true nearest neighbors and impostor nearest neighbors of each data point. Additionally, we plan to use the label information of training data to constrain the distances between pairwise data points in the same class. For example, we can add a penalty term using supervised stochastic neighbor embedding (SNE) \cite{SNE} or t-SNE \cite{tSNE} to constrain the within-class distances.

\section* {Acknowledgement}
We thank Geoff Hinton for his guidance and inspiration.  We thank Lee Zamparo for proofreading the manuscript and Jin Ke for 
drawing figure 1 and Figure 2.
\bibliographystyle{abbrv}
\bibliography{knndeepnet_min10}  

\end{document}